\documentclass[letterpaper,10pt,conference]{ieeeconf}
\IEEEoverridecommandlockouts
\usepackage[T1]{fontenc}
\usepackage{mathptmx}
\usepackage{graphicx}
\usepackage{amsmath,amssymb}
\usepackage{booktabs}
\usepackage[dvipsnames]{xcolor}

\makeatletter
\let\NAT@parse\undefined
\makeatother
\usepackage[
  colorlinks=true,
  linkcolor=BurntOrange,
  citecolor=BurntOrange,
  urlcolor=Orange
]{hyperref}
\newcommand{\robotname}{Pixel}
\newcommand{\SE}{\mathrm{SE}(3)}

\title{\LARGE\bf Pixel-UMI: Learning Environment-Aware Whole-Body\\
Manipulation from Sensorized Human Demonstrations}

\author{
Jason Jingzhou Liu$^{\ast}$, Tony Tao$^{\ast}$, Deepak Pathak$^{\dagger}$, Ruslan Salakhutdinov$^{\dagger}$\\[0.5em]
Carnegie Mellon University\\
$^{\ast}$Equal contribution\qquad $^{\dagger}$Equal advising
\\[0.4em]
\href{https://pixel-robot.com}{%
  \textbf{\texttt{https://pixel-robot.com}}%
}
}
\makeatletter
\newcommand{\pixelumiteaser}{%
\begin{minipage}{\textwidth}
  \centering
  \includegraphics[width=\linewidth]{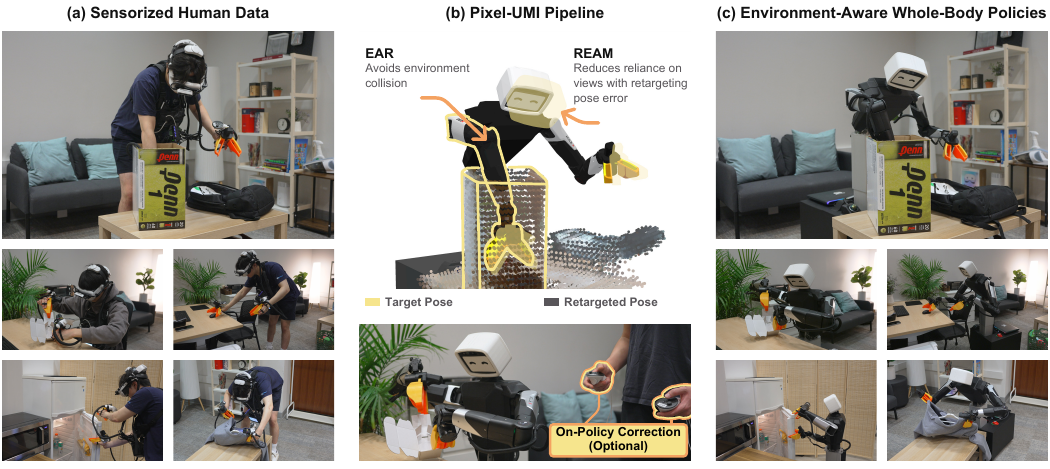}\par
  \nointerlineskip
  \setlength{\abovecaptionskip}{3pt}
  \def\@captype{figure}
  \caption{\textbf{Pixel-UMI overview.} (a) Humans collect whole-body mobile manipulation demonstrations with Pixel-UMI. (b) Environment-Aware Retargeting (EAR) uses offline scene reconstruction to convert human demonstrations into collision-aware whole-body trajectories for policy training. Retargeting-Error-Aware Masking (REAM) uses residual pose errors to reduce reliance on camera views the robot cannot reproduce accurately. Optional on-policy corrections further improve performance on contact-rich tasks. (c) Pixel-UMI enables long-horizon, contact-rich whole-body manipulation.}
  \label{fig:teaser}
  \vspace{0.5\baselineskip}
\end{minipage}%
}
\makeatother
\IEEEaftertitletext{\vspace{-12pt}\pixelumiteaser}
\begin{document}
\maketitle
\thispagestyle{plain}
\pagestyle{plain}
\begin{abstract}
Sensorized human demonstrations offer scalable data for mobile manipulation, but embodiment differences complicate policy learning. We present Pixel-UMI, a pipeline for learning environment-aware whole-body policies from human data. We first introduce Environment-Aware Retargeting (EAR), which combines reconstructed scene geometry with demonstrated head and gripper trajectories to determine robot posture and generate collision-aware joint and base targets. Delay-calibrated action assignment converts these trajectories into actions. However, kinematic differences can leave residual pose errors, creating viewpoint mismatches between human demonstrations and robot observations. We therefore introduce Retargeting-Error-Aware Masking (REAM), which uses these errors during training to reduce reliance on mismatched views at deployment. Finally, because estimated actions cannot fully capture contact-rich interactions, we collect on-policy human corrections through DAgger to improve performance. We evaluate Pixel-UMI on five challenging tasks spanning articulated-object manipulation, deformable-object handling, constrained-space reaching, whole-body collision avoidance, and contact-rich bimanual coordination, achieving 87.0\% average task progress. Video results are available at \url{https://pixel-robot.com/pixel-umi}
\end{abstract}
\enlargethispage{1\baselineskip}

\vspace{-1mm}
\section{Introduction}
\vspace{-1mm}
\begin{figure*}[!t]
 \centering
 \includegraphics[width=\textwidth]{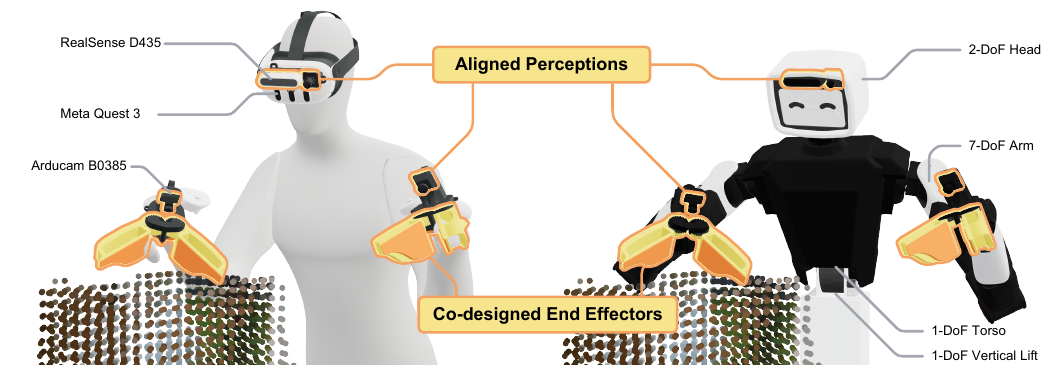}
 \caption{\textbf{Co-designed hardware for Pixel-UMI.} The sensorized human interface (left) and Pixel robot (right) use matched grippers and corresponding head and wrist camera layouts to align manipulation and visual observations.}
 \label{fig:hardware}
\end{figure*}

Mobile manipulation requires coordinated motion of the arms, torso, head, and base. Imitation learning provides a way to acquire these behaviors from demonstrations. Teleoperation is a widely used approach to collecting high-quality demonstrations~\cite{fu2024mobile,jiang2025behavior}, but is challenging for mobile manipulation: operators must coordinate many degrees of freedom, and collection across environments requires deploying a robot at each site.

Sensorized human data collection offers a more direct way to gather demonstrations by giving people a manipulation interface matched to the robot's~\cite{chi2024universal,sunday2025act1}. The Universal Manipulation Interface (UMI), for example, records images and motion through handheld grippers that the demonstrator uses to interact with objects. This allows people to coordinate their own bodies and receive physical feedback from contact, making demonstration more intuitive than teleoperation while allowing collection across environments without transporting a robot. Extensions to mobile manipulation complement the grippers with a head-mounted camera, providing the broader visual context needed for navigation and whole-body coordination~\cite{xu2026hommi,zeng2025activeumi}.
\enlargethispage{1\baselineskip}

Transferring these demonstrations to a robot requires aligning the demonstrated gripper and camera frames with their robot counterparts so that the manipulation and visual observations remain consistent across embodiments~\cite{chi2024universal,xu2026hommi}. Although task-frame matching can suffice for tabletop manipulation with limited redundancy, it leaves the posture of a whole-body mobile manipulator underdetermined: many arm, torso, and base configurations can realize similar gripper and head targets. These choices affect whether the robot can continue following the demonstrated motion within the scene. When retrieving a tennis-ball canister from a deep box, for example, the elbow must lift above the rim to allow the gripper to descend. Without an appropriate elbow posture, contact with the rim can obstruct the arm and prevent continued gripper tracking.

To support continued task-frame tracking, the robot must choose a body configuration that accommodates the surrounding geometry. To this end, we introduce Environment-Aware Retargeting (EAR), which uses scene geometry to resolve the robot's posture while matching the demonstrated gripper and head motion. EAR reconstructs static scene geometry from the episode's depth sequence, using temporal evidence to separate dynamic objects without requiring a separate scene scan. We then use environment-aware geometric fabrics~\cite{xie2020fabrics} to retarget the demonstrated motion against this geometry offline, producing body joint trajectories and base commands. The policy learns these posture choices directly, avoiding the need to combine scene estimation with collision-aware whole-body control during policy rollout.

However, fundamental differences in body shape and kinematics between humans and robots can still prevent exact head and gripper tracking, creating viewpoint mismatches between human training images and robot observations~\cite{xu2026hommi}. Retargeting alone cannot fully eliminate this mismatch, but we can mitigate its impact during policy training. We therefore introduce Retargeting-Error-Aware Masking (REAM). Because retargeting occurs before policy training, its pose errors reveal when a demonstrated viewpoint is difficult for the robot to reproduce. REAM uses these errors as geometric proxies for visual mismatch in percentile-based camera masking: views with higher-ranked errors are more likely to have their tokens excluded during training. This encourages the policy to draw on other available observations when a view is poorly aligned, while retaining access to all cameras at deployment.

Finally, sensorized human data collection without force or tactile sensors records motion states but not the interaction forces underlying human actions~\cite{chi2024universal,liu2025forcemimic,choi2026umift}. Motion states alone do not determine the robot commands needed to reproduce these interactions, so policies trained on this supervision can struggle with contact-rich manipulation. Nevertheless, learning from these demonstrations provides a useful policy prior for task motion and whole-body coordination, making on-policy correction practical. We therefore collect on-policy correction data through DAgger~\cite{ross2011dagger} to improve performance on contact-rich whole-body tasks.

We call our combined pipeline Pixel-UMI and evaluate it on the co-designed \robotname{} platform~\cite{pixel2027system}. We summarize our key contributions as follows:
\begin{itemize}
 \interlinepenalty=10000
 \item EAR converts head and gripper demonstrations into environment-aware whole-body joint and base supervision. Across three tasks requiring whole-body coordination, it increases average task progress by 64.41~pp.
 \item REAM uses percentile-ranked retargeting errors to guide camera masking and reduce dependence on mismatched views. On the same three tasks, adding REAM improves average task progress by 14.05~pp.
 \item We evaluate on-policy correction through DAgger to address limitations of state-based UMI supervision. On two challenging, long-horizon, contact-rich bimanual whole-body tasks, two DAgger rounds improve average task progress by 72.95~pp.
\end{itemize}

\begin{figure*}[!t]
 \centering
 \includegraphics[width=\textwidth]{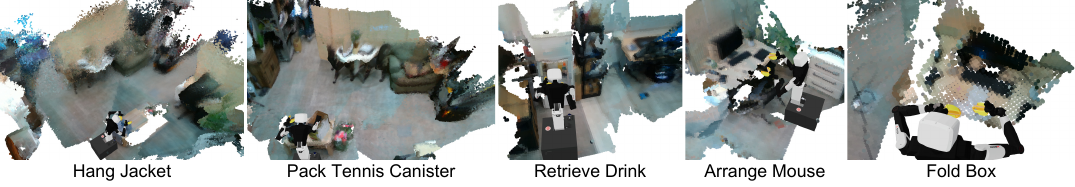}
 \caption{\textbf{Static scene reconstructions.} Episode-level static geometry reconstructed from sensorized human demonstrations across five tasks, shown with the retargeted robot. These point clouds provide the environment geometry for downstream collision-aware retargeting.}
 \label{fig:static-reconstructions}
\end{figure*}

\section{Hardware}
Pixel-UMI uses the co-designed robot and sensorized human interface of \robotname{} (Figure~\ref{fig:hardware}). The human-worn grippers match the robot grippers in manipulation geometry, visual appearance, and wrist-camera placement. Together with aligned head RGB cameras, this co-design reduces the physical and visual gaps between embodiments to support the conversion of sensorized human demonstrations into high-quality robot training data. Each human-worn gripper is equipped with an encoder to record gripper width. Meta Quest controllers track the gripper poses relative to the head, while a Meta Quest 3 headset worn by the demonstrator tracks the head pose in the world frame. We refer readers to the Pixel system paper~\cite{pixel2027system} for further hardware details.

We additionally equip the human headset with a depth camera to record metric depth throughout each demonstration. We use these observations for offline metric 3D reconstruction to support collision-aware retargeting from human demonstrations to robot demonstrations. We note that during robot policy rollout, visual observations come from the head and wrist RGB cameras; the robot-side depth camera is not used.

\section{Pixel-UMI}
\label{sec:method}

We present Pixel-UMI, a pipeline for learning whole-body robot policies from sensorized human demonstrations. Our goal is to address the fundamental human-to-robot gaps involved in converting these demonstrations into high-quality robot training data.

Firstly, the recorded head and gripper poses leave the robot's whole-body posture underdetermined due to kinematic redundancy. To address this, we introduce Environment-Aware Retargeting (EAR). We reconstruct scene geometry offline and use collision avoidance to guide the robot's whole-body posture while tracking the demonstrated motion. However, the resulting trajectories still only describe robot states, not robot actions. We therefore assign pseudo-actions using calibrated joint tracking delays, so that the training targets better account for the robot's response.

Secondly, differences in human and robot kinematics and body geometry can still prevent exact tracking. As a result, the head and wrist camera views during robot policy rollout can differ from those recorded in the human demonstrations. To alleviate this visual gap, we introduce Retargeting-Error-Aware Masking (REAM), which uses retargeting errors during training to encourage the policy to rely less on poorly matched camera views.

Finally, remaining human-to-robot gaps can still limit policy performance on long-horizon, contact-rich manipulation tasks. We therefore collect a small amount of on-policy correction data through DAgger to improve policy performance on these tasks.

\subsection{Preliminaries}
\label{sec:preliminaries}

For a synchronized human demonstration, let $I_t^c$ denote the RGB image from camera $c\in\{H,L,R\}$, where $H$, $L$, and $R$ denote the head, left gripper, and right gripper, respectively, and let $D_t$ denote the synchronized head-camera depth image. We use the same indices for task frames, with $T_t^j\in\SE$ denoting the pose of frame $j\in\{H,L,R\}$ in a common world frame, obtained from the Meta Quest 3 headset and controller tracking. The left and right gripper states $g_t^L,g_t^R$ are derived from their encoder readings. An episode contains $T$ such samples and a task instruction $\ell$.

We represent the robot configuration as $x_t=(b_t,q_t)$, where $b_t=(x_t^b,y_t^b,\psi_t)$ is the planar base pose and $q_t\in\mathbb{R}^{20}$ contains lift, torso, arm, head, and gripper coordinates. The learned action is
\begin{equation}
 a_t=[v_{x,t}^{B},v_{y,t}^{B},\omega_t,q_t^{\mathrm{cmd}}]
 \in\mathbb{R}^{23},
 \label{eq:action}
\end{equation}
with planar velocities expressed in the robot base frame and absolute position targets for the body and grippers. Thus, the policy directly specifies joint-space targets, eliminating the need for an online whole-body controller (WBC) to choose motion in the task-space null space.

\subsection{Environment-Aware Retargeting (EAR)}
\label{sec:retarget}

\subsubsection{\textbf{Metric reconstruction from the demonstration}}
Our goal is to reconstruct a metric 3D scene for each collected episode, providing the environment geometry for obstacle avoidance during retargeting. The Meta Quest 3 tracking system provides head and gripper trajectories in a common world frame, with synchronized depth images from the head-mounted camera at each timestep. However, potential pose drift or tracking discontinuities can cause depth point clouds to misalign when combined using the recorded poses alone. To address this, we perform offline point-cloud registration and pose refinement, using geometric alignment across observations to refine the head trajectory before fusing the corrected point clouds into a scene. The reconstruction uses observations from the full recording, including a brief floor-calibration observation when needed; a separate scan of the workspace is not required.

We convert each depth image into a metric point cloud in the world frame using the calibrated camera intrinsics, head-to-camera transform, and recorded head pose. Before registration, we filter points near the cameras and tracked grippers to reduce the influence of the demonstrator and nearby manipulated objects. We then identify candidate static surfaces through temporal persistence: after voxelizing the clouds in the recorded world frame, we retain points whose voxels are observed in multiple frames within a local temporal window. This favors surfaces that remain at approximately the same world location across nearby observations. We downsample these candidates and select keyframes to balance temporal coverage, camera motion, and geometric overlap.

We align overlapping keyframes using point-to-plane iterative closest point (ICP) registration~\cite{besl1992registration,chen1992object}. To limit unreliable registration constraints, we discard alignments with insufficient overlap or too few matched points, large distances between matched points, or excessive pose corrections relative to the VR initialization. We then refine the camera trajectory through pose-graph optimization~\cite{choi2015robust}, combining the accepted ICP constraints with relative-motion priors from the recorded VR poses of consecutive keyframes. We identify possible tracking discontinuities as abrupt changes in vertical or total camera translation that exceed displacement and velocity thresholds and are outliers relative to the episode's motion, measured using a median-based outlier score. For these intervals, we reduce the weights on the corresponding VR relative-motion residuals in the pose-graph objective, allowing geometric alignment to have greater influence on the refined trajectory. We interpolate the resulting pose corrections to all frames.

Finally, we establish a ground-referenced world frame for retargeting. EAR solves for absolute whole-body joint configurations with the mobile base supported by the floor, so the head and gripper targets must be expressed at the correct metric heights above that floor. An arbitrary vertical offset in the default VR fixed frame would otherwise shift these targets relative to the robot's base and change the resulting body posture. We estimate the ground plane from observed floor points and vertically shift the head and gripper trajectories and reconstructed scene together so that ground level corresponds to $z=0$ in the fixed retargeting frame. For tasks whose demonstrations may not reveal the floor, such as seated box folding at a table, we require the demonstrator to look down and record a brief floor observation before beginning the task.

\subsubsection{\textbf{Separating static structure from dynamic geometry}}
Given the reconstructed scene, we remove the objects being manipulated so that collision avoidance does not prevent the contact required by the task. We make the simplifying assumption that these objects move or deform during the demonstration, and use this motion to separate them from static environmental structure.

We compare aligned depth observations across the episode using occupancy and free-space evidence~\cite{hornung2013octomap}. Each depth measurement indicates a surface at its endpoint and empty space along the viewing ray before it. If another frame sees through a previously occupied location to a farther surface, this provides evidence that the earlier surface moved~\cite{schauer2018peopleremover}. We retain candidate voxels as static when they have repeated surface observations across time and few conflicting free-space observations, excluding transient geometry and regions with insufficient evidence. Objects whose motion is not observed may remain in the static map. Figure~\ref{fig:static-reconstructions} shows the resulting static geometry used for collision-aware retargeting.

\subsubsection{\textbf{Collision-aware whole-body retargeting}}
\begin{figure}[!t]
 \centering
 \includegraphics[width=\columnwidth]{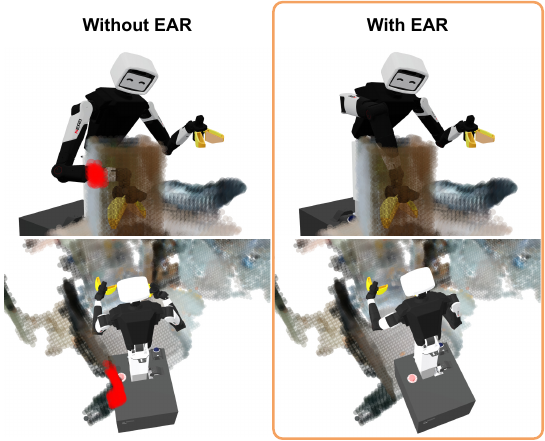}
 \caption{\textbf{Effect of collision-aware retargeting.} Without EAR (left), retargeted robot postures intersect the surrounding scene, with collisions highlighted in red. EAR (right) uses reconstructed geometry to select whole-body postures that avoid these obstacles and support continued tracking of the demonstrated head and gripper motion.}
 \label{fig:ear-comparison}
\end{figure}

\begin{figure*}[!t]
 \centering
 \includegraphics[width=\textwidth]{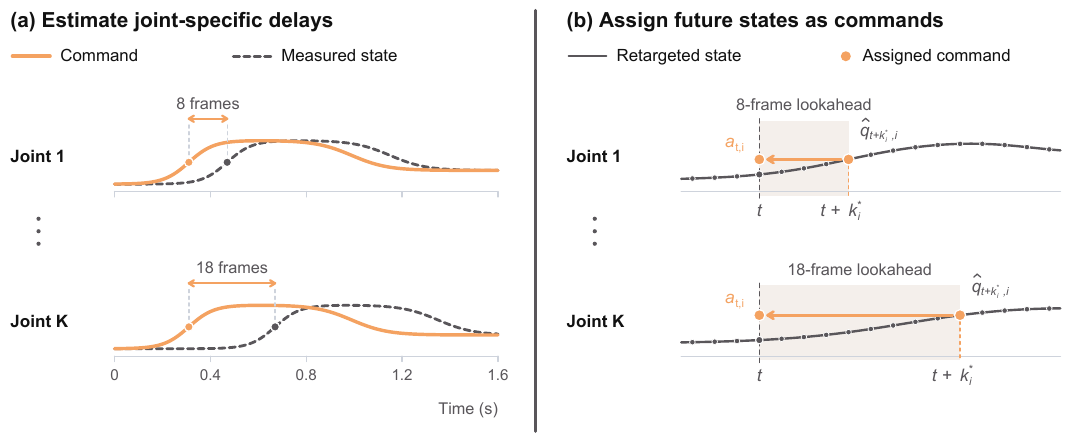}
 \caption{\textbf{Dynamics-aware action assignment.} (a) Joint-specific delays are estimated by aligning robot commands and measured states. (b) These delays determine how far ahead to sample the retargeted state when constructing each joint's command label.}
 \label{fig:action-assignment}
\end{figure*}

Given the static scene represented as a voxelized point cloud and the corrected head and gripper trajectories, our goal is to solve for robot joint and base configurations that track the demonstrated poses while avoiding the surrounding geometry. We use geometric fabrics~\cite{xie2020fabrics} as an offline retargeting solver. This framework composes motion objectives defined in different task spaces through metric-weighted acceleration updates, allowing tracking, collision avoidance, and joint-limit avoidance to jointly shape whole-body motion. Our formulation combines head and gripper pose attraction, environment and self-collision repulsion, joint-limit repulsion, nominal-posture regularization, and mobile-base and joint-priority terms.

We represent the robot's collision geometry with spheres attached to its links. For a sphere with center $c_s(x)$ and radius $r_s$, its clearance from the static scene $\mathcal S$ is
\begin{equation}
 d_s(x,\mathcal S)=\min_{p\in\mathcal S}\|c_s(x)-p\|_2-r_s.
 \label{eq:clearance}
\end{equation}
As this clearance decreases, repulsive fabric terms increasingly influence the robot's posture, including links whose motion is not specified by the head and gripper targets. We solve sequentially along the demonstrated trajectory, using the preceding robot configuration to initialize each step. Figure~\ref{fig:ear-comparison} illustrates how incorporating scene geometry changes the whole-body posture to support continued task-frame tracking.

In our formulation, collision avoidance acts as a soft constraint, with repulsion balanced against pose tracking through the fabric metrics. This differs from QP-based differential inverse kinematics formulations such as Mink~\cite{zakka2026mink} and Pink~\cite{caron2026pink}, whose collision-avoidance limits and barriers impose hard inequalities in the QP. We choose this soft formulation to accommodate reconstruction noise, such as isolated floating points: these points contribute repulsion without becoming inviolable geometric boundaries. The solver can therefore trade off tracking and clearance locally when the reconstructed geometry contains spurious obstacles.

\subsubsection{\textbf{Dynamics-aware action assignment}}
\label{sec:assignment}
Retargeting produces whole-body state trajectories, but behavior cloning requires robot actions. We first time-scale the trajectories to reduce motion speed and resample their associated images. Sunday likewise reports using slower human demonstrations and policy rollouts~\cite{sunday2026technology}. We then derive base velocities by differentiating the base poses and expressing them in the robot frame. For body joints, next-state action labels ignore differences in how quickly each joint tracks its target. We therefore calibrate a tracking delay for each joint and use it to construct pseudo-action labels, illustrated in Figure~\ref{fig:action-assignment}.

\begin{figure*}[!t]
 \centering
 \includegraphics[width=\textwidth]{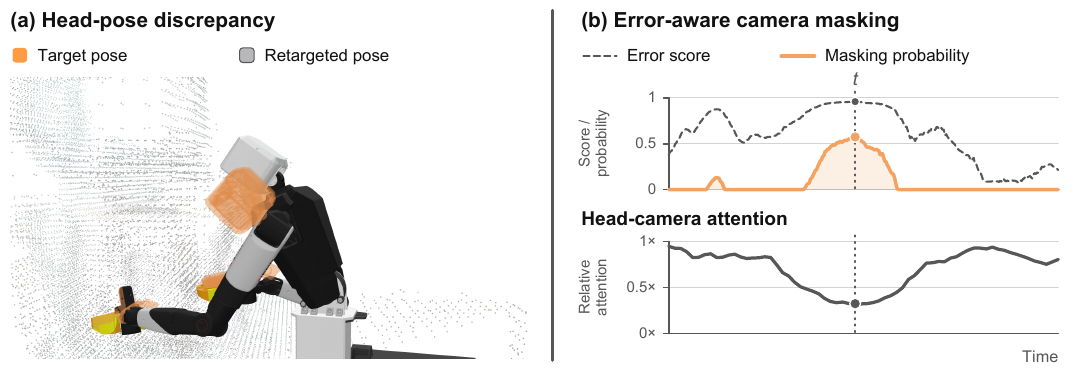}
 \caption{\textbf{Retargeting-Error-Aware Masking (REAM).} (a) Head-pose mismatch between the demonstrated target (orange) and retargeted pose (gray). (b) Retargeting error and masking probability (top), with measured relative head-camera attention (bottom), for a transformed Retrieve Drink demonstration used for training. Attention decreases during the high-error interval.}
 \label{fig:ream}
\end{figure*}

We estimate these delays using actual robot trajectories, such as teleoperation data, with recorded joint positions and corresponding position targets. For joint $i$, let $a_{t,i}^{n}$ denote the target issued at time $t$ in trajectory $n$, and $q_{t,i}^{n}$ the measured position. We select the offset that best aligns each target with the state reached later:
\begin{equation}
 k_i^* = \underset{k\in\{0,\ldots,K\}}{\operatorname{argmin}}
 \sum_n\sum_t
 \left(q_{t+k,i}^{n}-a_{t,i}^{n}\right)^2,
 \label{eq:lagcalibration}
\end{equation}
where we sum over trajectories and timestamps with valid future states for all candidate offsets. Calibration trajectories need not be paired with human demonstrations.

For the retargeted joint trajectory $\hat q_t$, we assign
\begin{equation}
 a_{t,i}=\hat q_{t+k_i^*,i},
 \label{eq:assignment}
\end{equation}
advancing each joint's target by its calibrated delay. UMI also compensates for deployment latency~\cite{chi2024universal}; here, we construct per-joint action labels for whole-body trajectories.

We highlight that this approximation accounts for tracking lag in free-space motion, but contact can prevent a joint from reaching its target. A target beyond a contact surface may generate force without further motion, so the state trajectory alone cannot determine the appropriate action. We address this gap through on-policy correction (Section~\ref{sec:dagger}).

The resulting joint targets and base velocities form the action representation in Equation~\ref{eq:action}. The policy learns scene-dependent whole-body posture directly and sends these actions to the robot's low-level controllers during rollout, without requiring online scene reconstruction or task-space retargeting.

\subsection{Retargeting-Error-Aware Masking (REAM)}
\label{sec:masking}

Differences in human and robot kinematics and body geometry can prevent exact tracking of the demonstrated head and gripper poses. During drink retrieval, for example, the human can lean and reorient the head in ways that the robot cannot reproduce, illustrated in Figure~\ref{fig:ream}. Consequently, the human images used to train the policy can differ from the views available during robot rollout. A policy that relies on these mismatched views may therefore transfer poorly.

We introduce REAM to alleviate this visual gap during policy training. Because offline retargeting provides both the demonstrated and retargeted poses throughout each episode, we can identify when and where tracking errors occur. We use these errors to increase the probability of masking visual tokens from poorly aligned cameras, encouraging the policy to rely less on those views when predicting actions.

For a given camera and timestep, we omit the camera and time indices for clarity. Let $(p,R)$ and $(\hat p,\hat R)$ denote the demonstrated and retargeted positions and orientations of the associated head or gripper frame. We measure their discrepancy as
\begin{equation}
 e_p = \|p-\hat p\|_2, \qquad
 e_R = \left\|\operatorname{Log}(R\hat R^\top)^\vee\right\|_2,
 \label{eq:poseerror}
\end{equation}
where $e_p$ is the position error and $e_R$ is the rotation error in radians. Here, $\operatorname{Log}$ denotes the matrix logarithm on $\mathrm{SO}(3)$, and $(\cdot)^\vee$ maps a skew-symmetric matrix to its corresponding rotation vector. These errors serve as geometric proxies for camera viewpoint mismatch.

We compare each error with the errors observed in the training set. We pool position errors across cameras and timesteps into a sorted list, and construct a separate list for orientation errors. Let $F_p(e)$ and $F_R(e)$ return the percentile rank of error $e$ in these lists, normalized to $[0,1]$. Both mappings are computed from the training data and shared across cameras.

We average the two ranks to obtain a combined error score $r\in[0,1]$:
\begin{equation}
 r=\frac{F_p(e_p)+F_R(e_R)}{2}.
 \label{eq:errorscore}
\end{equation}
We map this score to a masking probability:
\begin{equation}
 \pi=
 \begin{cases}
  0, & r\leq\rho,\\[4pt]
  p_{\max}\left(\dfrac{r-\rho}{1-\rho}\right)^\gamma, & r>\rho.
 \end{cases}
 \label{eq:percentilemask}
\end{equation}
The threshold $\rho\in[0,1)$ determines when masking begins. Above this threshold, the probability increases with the error score, reaching $p_{\max}$ at $r=1$; $\gamma>0$ controls the shape of this increase.

We apply this probability when sampling camera masks for the flow-matching policy described in the \robotname{} system paper~\cite{pixel2027system}. For each training sample, let $\pi_{t,c}$ denote the masking probability for camera $c$ at observation time $t$. We draw a Bernoulli mask for each camera using this probability. A masked camera has all its visual tokens across the observation window excluded as attention keys and values. For cross-attention, we implement this with an additive mask:
\begin{equation}
 \operatorname{Attn}(Q,K,V;M_t)
 =\operatorname{softmax}\!\left(\frac{QK^\top}{\sqrt d}+M_t\right)V,
 \label{eq:maskedattention}
\end{equation}
where $Q$ contains action-token queries, $K$ and $V$ contain context keys and values, and $d$ is the key dimension. We set columns of $M_t$ associated with a masked camera to $-\infty$, making their attention weights zero; all other entries are zero. The action targets used for supervision remain unchanged.

Figure~\ref{fig:ream} shows the retargeting error, masking probability, and measured relative head-camera attention for a transformed Retrieve Drink demonstration from the training set. Higher masking probabilities coincide with lower relative attention to the head camera. This masking encourages the policy to predict actions with less reliance on views that may become out of distribution during rollout due to camera-pose mismatch. During rollout, we retain all views without using human reference trajectories or online retargeting errors.

\subsection{On-Policy Correction via DAgger}
\label{sec:dagger}

A policy trained on EAR-transformed demonstrations with REAM can already achieve strong performance on various tasks, but may still struggle with challenging long-horizon, contact-rich manipulation such as box folding and jacket hanging, illustrated in Figure~\ref{fig:all-tasks}. EAR and REAM reduce the human-to-robot gap, but our sensorized human interface records states rather than actions. These demonstrations therefore cannot fully specify the robot actions needed for complex contact interactions.

Nevertheless, these demonstrations provide useful motion priors. The policy already produces approximately the desired motion, making it easier to provide targeted action corrections than to teleoperate the entire task without policy assistance. We use this policy to collect on-policy corrections via DAgger~\cite{ross2011dagger,kelly2019hgdagger,hu2025rac}.

Our policy predicts joint-space actions, whereas the VR controllers provide relative end-effector corrections. We apply these corrections to the robot's current end-effector poses and pass the resulting targets to the same WBC used for teleoperation in the \robotname{} system paper~\cite{pixel2027system}. The WBC converts these targets into joint-space actions while using the policy's joint space predictions for posture regularization.

To support a smooth return to policy rollout after an intervention, we use real-time chunking (RTC)~\cite{black2025training}. We retain a short history of the corrected actions and supply its final steps as a fixed prefix for the next policy prediction, with the prefix length determined by the estimated inference delay. The flow-matching sampler holds this prefix fixed while generating the remainder of the action chunk. We resume rollout from the generated suffix, conditioning the continuation on the recent corrected motion.

We record complete rollout trajectories, including policy-generated actions and human corrections. After each round, we aggregate these robot observation--action pairs with the processed human demonstrations and robot data from earlier rounds, and train a new policy from scratch.

\section{Experiments}
\label{sec:experiments}
We evaluate Pixel-UMI on five mobile manipulation tasks to address four questions:

\begin{enumerate}
    \item Does EAR improve task performance compared with retargeting that does not account for the environment?
    \item Which action representation is most effective, and does delay-calibrated action assignment improve policy performance?
    \item Which observation space is most effective, and can REAM improve transfer in the presence of retargeting errors?
    \item How can we effectively use on-policy correction data collected through DAgger?
\end{enumerate}

We study the first three questions on three tasks using policies trained only on human demonstrations. For the fourth, we use two additional contact-rich tasks to evaluate learning from on-policy correction data.

\subsection{Tasks and Evaluation Metric} 
\label{sec:tasks}

Figure~\ref{fig:all-tasks} illustrates the human demonstrations and policy rollouts for all the tasks. The demonstration counts listed below refer to human demonstrations.

\newsavebox{\taskfigurecaptionbox}
\begin{figure*}[p]
    \centering
    \setbox\taskfigurecaptionbox=\vbox{%
        \hsize=\textwidth
        \caption{We evaluate EAR, REAM, and Action Assignment across three challenging mobile manipulation tasks and evaluate on-policy correction on a further two tasks. Video results are available at \url{https://pixel-robot.com/pixel-umi}}
        \label{fig:all-tasks}
    }
    \includegraphics[
        height=\dimexpr\textheight-\ht\taskfigurecaptionbox-\dp\taskfigurecaptionbox-1pt\relax
    ]{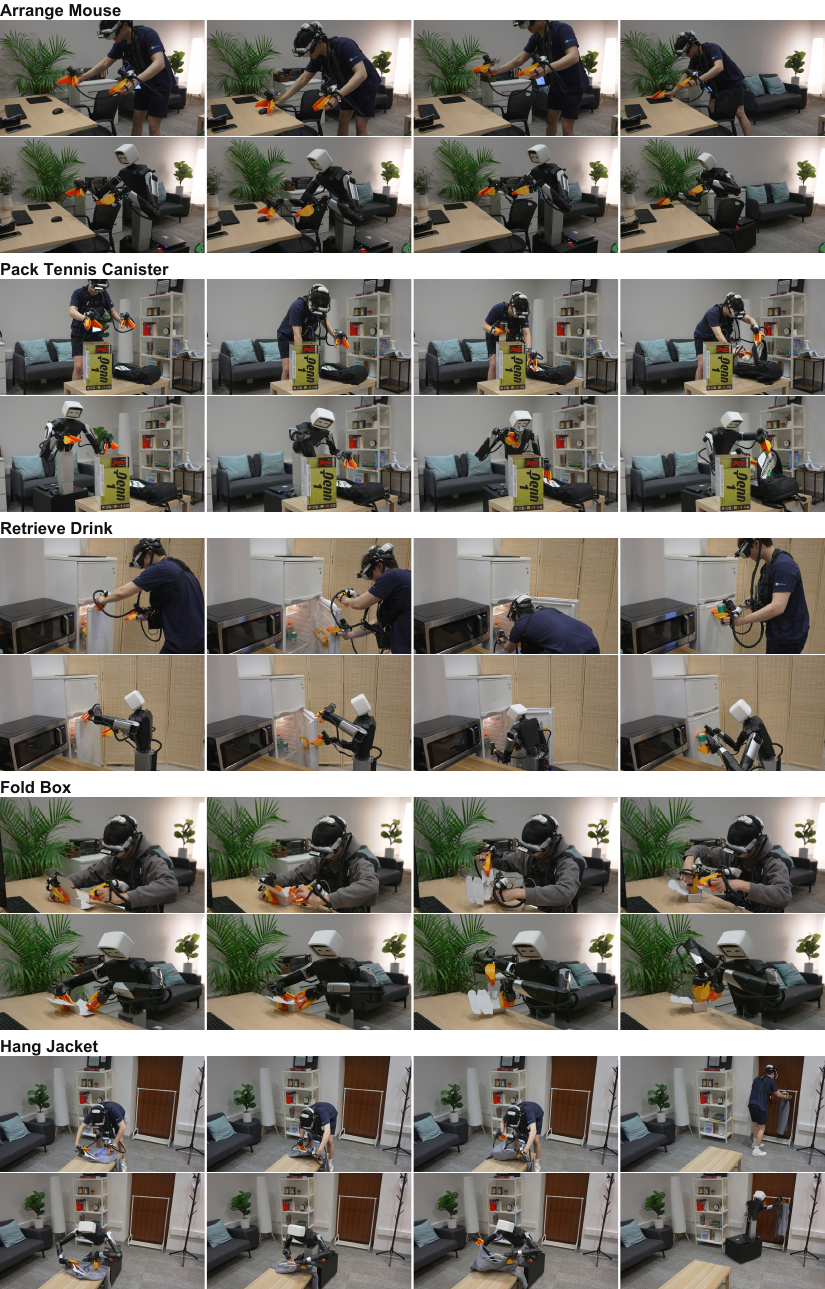}\par
    \nointerlineskip
    \box\taskfigurecaptionbox
\end{figure*}

\begin{enumerate}
    \item \textbf{Arrange Mouse (100 demonstrations).}
    The robot must retrieve a computer mouse and place it on a mouse pad on a desk while avoiding a chair positioned in front of the desk. This task evaluates whole-body coordination and collision avoidance while moving through cluttered environments, requiring the robot to dynamically adjust its posture during retrieval and placement.

    \item \textbf{Pack Tennis Canister (178 demonstrations).}
    The robot must reach into a deep cardboard box to retrieve a tennis-ball canister, open a bag, and place the canister inside. This task evaluates constrained-space reaching and bimanual manipulation, requiring the robot to maneuver its elbow around the box rim while reaching inside.

    \item \textbf{Retrieve Drink (219 demonstrations).}
    The robot must open a refrigerator, reach inside to retrieve a drink can, and close the door while avoiding nearby shelving. This task evaluates articulated-object manipulation and robustness through mismatches in head viewpoint.

    \item \textbf{Fold Box (393 demonstrations).}
    The robot must fold a cardboard blank into a box, insert the side tabs, and crimp the edges to secure it. This task evaluates precise, contact-rich bimanual manipulation through a sequence of folding, insertion, and crimping operations.

    \item \textbf{Hang Jacket (261 demonstrations).}
    The robot must pick up a hanger, insert it into a jacket, wrap the garment around the hanger, and hang it on a rack. This task evaluates deformable-object manipulation, bimanual coordination, and grasp transitions across multiple stages of manipulation.
\end{enumerate}

For each experiment, we conduct 20 evaluation trials across four randomized initial configurations. We report \emph{task progress} on a scale from 0\% to 100\%, with discrete task-completion stages weighted according to their difficulty. 

\subsection{Policy Details}
We use single-task policies for all ablation experiments.
For Arrange Mouse, Pack Tennis Canister, and Retrieve Drink, we train policies using only human demonstrations. For Fold Box and Hang Jacket, we first train on human demonstrations, then add 60 successful on-policy robot correction demonstrations per task in each of two rounds.

The policy uses pretrained DINOv3~\cite{simeoni2025dinov3} ViT-B/16 visual backbones and a flow-matching~\cite{lipman2023flow} action decoder, following the \robotname{} system paper~\cite{pixel2027system}. Unless otherwise stated, the policy receives synchronized RGB images from the head camera and both wrist cameras, along with robot joint states. Head images are processed at $384\times384$ and wrist images at $224\times224$.

We initialize the action decoder from scratch and jointly fine-tune the pretrained visual backbones for 150,000 optimizer updates.

The policy predicts a chunk of 50 actions spanning one second at a control rate of $50\,\mathrm{Hz}$. Each 23-dimensional action contains 20 absolute position targets for the lift, torso, arms, head, and grippers, together with the planar base velocities $(v_x^B,v_y^B,\omega)$ expressed in the robot base frame. These commands are sent directly to the low-level controllers during rollout, without an online task-space whole-body controller.

We use real-time chunking (RTC)~\cite{black2025training} as in the system paper. This compensates for inference latency and smooths transitions between consecutive action chunks.

\subsection{Environment Awareness}

EAR optimizes robot motion offline to avoid collisions with the environment. We evaluate whether this supervision leads to collision avoidance during policy rollout and improves task performance. We choose three tasks that impose different geometric constraints. Pack Tennis Canister requires an arm configuration that keeps the elbow clear of the box rim while the gripper reaches inside. Arrange Mouse requires whole-body coordination during motion, as the robot must dynamically adjust its posture to avoid the chair while retrieving and placing the mouse. Retrieve Drink requires the arms to reach inside the refrigerator without collision while the base remains clear of the adjacent cabinet. We evaluate the following two experimental conditions:

\noindent\textbf{EAR only.} We use trajectories retargeted with environment collision avoidance enabled.

\noindent\textbf{No environment awareness.} We use trajectories retargeted without environment collision avoidance.

Both conditions retain self-collision avoidance and use absolute joint-space control with the full observation space described above. REAM is disabled in both conditions.

\subsection{Action Representation and Assignment}

We first compare three action representations on Arrange Mouse, Pack Tennis Canister, and Retrieve Drink to assess how the action space affects transfer of EAR's whole-body motion.

\noindent\textbf{Joint Abs.} The policy predicts absolute joint-position targets together with planar base velocities, directly specifying the robot's whole-body posture.

\noindent\textbf{Joint Rel.} The policy expresses joint targets as relative offsets while retaining the same base-velocity commands.

\noindent\textbf{EEF-Rel Base.} Following the relative-base representation in the \robotname{} system paper~\cite{pixel2027system}, the policy predicts head and bimanual end-effector target poses in a shared, ground-referenced pseudo-base frame for each action chunk. We retain joint angles for the gripper. An online whole-body controller converts these targets into robot motion. This tests whether task-space targets suffice without directly learning EAR's posture choices.

Based on this comparison, we select \textit{Joint Abs} and test whether accounting for joint response delays improves performance. We hold the retargeted state trajectories fixed and compare two action-labeling schemes:

\noindent\textbf{Calibrated labels.} We assign joint targets using per-joint lookahead settings calibrated from previously collected robot teleoperation data, following Section~\ref{sec:assignment}.

\noindent\textbf{Next-state labels.} We use the next retargeted joint state as the action, providing a reference without per-joint delay calibration.

All action-representation and assignment conditions use EAR and the full observation space, with REAM disabled.

\subsection{Observation Space and Retargeting Error Aware Masking}

Retargeting cannot always reproduce the demonstrated head and wrist poses exactly, creating viewpoint mismatches between human training images and robot observations at deployment. We investigate how camera observations and head-tracking objectives affect policy performance on Arrange Mouse, Pack Tennis Canister, and Retrieve Drink. We compare four conditions:

\noindent\textbf{All cameras (EAR only).} We use standard head and wrist retargeting and train with all three camera views, with REAM disabled.

\noindent\textbf{Wrist only.} We retarget only the wrist poses and use only the two wrist cameras during training and deployment, with REAM disabled. This tests whether removing the head view and its tracking objective improves transfer despite the loss of visual context.

\noindent\textbf{Relaxed head tracking.} We reduce the head-pose tracking weight during retargeting while retaining all three cameras, with REAM disabled. This tests whether favoring wrist tracking at the expense of head alignment improves transfer.

\noindent\textbf{REAM.} We retain standard retargeting and all three cameras, but mask camera observations during training according to their retargeting pose errors. This tests whether error-guided masking can mitigate viewpoint mismatch while preserving visual context at deployment.

All conditions use EAR, \textit{Joint Abs} actions, and calibrated action assignment. For REAM, we use an 80th-percentile cutoff ($\rho=0.8$) and a maximum masking probability of 90\% ($p_{\max}=0.9$) in Equation~\ref{eq:percentilemask}. The masking probability is zero at or below the cutoff and increases quadratically for higher error-rank scores. Masking is applied only during training; all three cameras remain available at deployment.

\subsection{On-Policy Correction}

We evaluate whether DAgger~\cite{ross2011dagger} improves policies trained on human demonstrations and which parts of the collected robot data are most useful. We study Fold Box and Hang Jacket, which require long-horizon, contact-rich bimanual manipulation.

Starting from a human-trained policy, an operator corrects rollout errors using the procedure in Section~\ref{sec:dagger}. We collect 60 successful corrected robot rollouts per task in each of two rounds, for a total of 120 per task. We compare three training conditions:

\noindent\textbf{Full rollouts.} We combine human demonstrations with complete corrected robot rollouts, including both policy-executed and operator-controlled segments.

\noindent\textbf{Interventions only.} We combine human demonstrations with only the operator-controlled segments of the same rollouts. Comparison with \textit{Full rollouts} tests the value of retaining policy-executed segments alongside corrections.

\noindent\textbf{Robot-only demos.} We train on all complete robot rollouts from both rounds without human demonstrations, testing whether human data remain useful once robot correction data are available.

After each round, we retrain \textit{Full rollouts} and \textit{Interventions only} from scratch on the original human dataset plus their respective robot data accumulated across all completed rounds. The policy trained after round one is used to collect round two. All conditions use uniform sampling.

\section{Results}
\label{sec:results}

Videos of results are available at \url{https://pixel-robot.com/pixel-umi}

\begin{figure*}[t]
 \centering
 \includegraphics[width=0.888141027\textwidth]{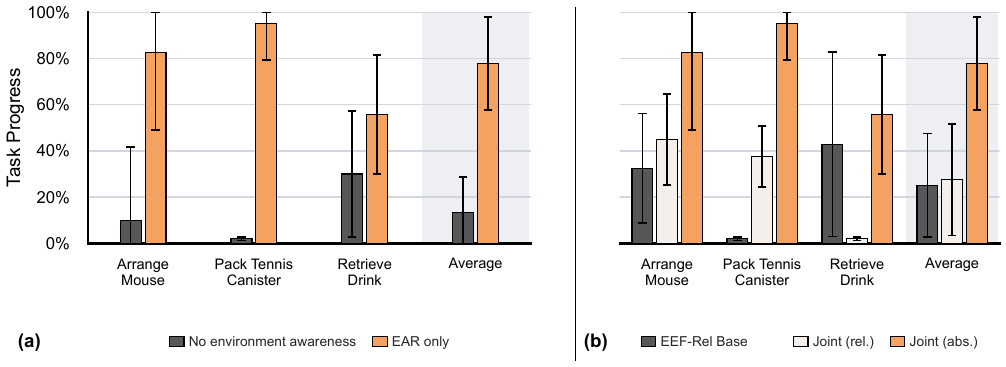}
 \caption{\textbf{EAR and action-representation ablations.} (a) EAR only versus retargeting without environment awareness. (b) EEF-Rel Base versus Joint Rel and Joint Abs. Both panels omit REAM.}
 \label{fig:retarget-results}
\end{figure*}

\begin{figure*}[t]
 \centering
 \includegraphics[width=\textwidth]{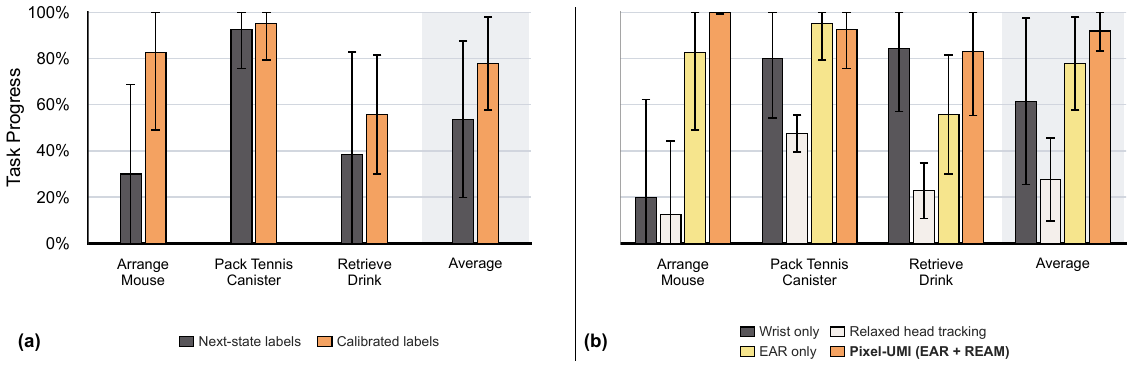}
 \caption{\textbf{Action-assignment and REAM ablations.} (a) Robot-calibrated command labels versus next-state labels, with REAM omitted. (b) Pixel-UMI (EAR + REAM) versus EAR only, wrist-only observations, and relaxed head tracking.}
 \label{fig:observation-results}
\end{figure*}

\subsection{EAR Improves Whole-Body Manipulation}
As shown in Figure~\ref{fig:retarget-results}, EAR improves task progress across all three tasks, increasing average progress by 64.41 percentage points over retargeting without environment awareness. The largest gains occur in Arrange Mouse and Pack Tennis Canister, where nearby obstacles constrain the body configurations needed to complete the task.

In Pack Tennis Canister, the policy trained without environment awareness hits the box rim with its elbow as the gripper descends. This collision prevents the robot from reaching the canister, resulting in zero task progress. EAR instead supplies training trajectories that keep the elbow clear of the rim while allowing the gripper to reach into the box. In Arrange Mouse, the policy trained without environment awareness often collides with the chair, either while approaching the mouse or after grasping it and moving toward the mouse pad. Avoiding these collisions requires coordinating the robot's body throughout both retrieval and placement. EAR also improves Retrieve Drink, where the robot must reach into the refrigerator while remaining clear of nearby shelving and cabinets.

The qualitative comparison in Figure~\ref{fig:ear-comparison} illustrates how incorporating scene geometry during retargeting changes the robot's whole-body posture. These scene-aware configurations provide the policies with a strong prior for moving around obstacles while continuing the demonstrated manipulation. The results support using environment geometry during offline retargeting to teach policies the coordinated body motion needed for manipulation in constrained spaces.

\subsection{Absolute Joint Targets and Delay Calibration Improve Policy Performance}
With EAR enabled, Joint Abs outperforms EEF-Rel Base and Joint Rel across all three tasks in Figure~\ref{fig:retarget-results}, achieving 77.74\% average task progress compared with 25.12\% and 27.50\%, respectively. EEF-Rel Base specifies head and end-effector poses but leaves the remaining body configuration to an online whole-body controller. These targets underspecify the posture needed to avoid nearby obstacles, and the controller often produces motions that collide with the environment. In Pack Tennis Canister, this representation fails to preserve the elbow configuration needed to reach inside the box, resulting in zero task progress.

\begin{figure*}[t]
 \centering
 \includegraphics[width=\textwidth]{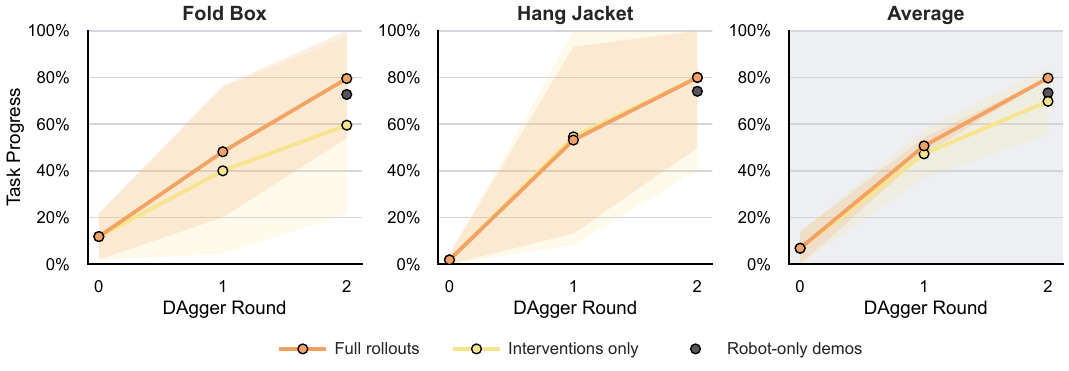}
 \caption{\textbf{Policy improvement through DAgger.} We compare training on complete corrected rollouts with intervention segments only. Gray markers show policies trained on robot demonstrations alone.}
 \label{fig:dagger-results}
\end{figure*}

\begingroup
\widowpenalty=10000
Joint Rel retains joint-level supervision from EAR's environment-aware trajectories. However, errors accumulate across relative commands, causing the robot to drift away from the demonstrated trajectories and quickly encounter states outside the training distribution. This is particularly problematic for tasks requiring precise positioning. In Retrieve Drink, the gripper must enter a small notch in the refrigerator door to open it. With Joint Rel, accumulated errors often cause the gripper to miss this notch, preventing further progress. Joint Abs directly specifies the target joint configuration at each step, helping preserve both the demonstrated posture and the precision needed for these interactions.
\par\endgroup

Holding Joint Abs fixed, we next compare calibrated action assignment with next-state labels. Figure~\ref{fig:observation-results} shows that calibration increases average task progress by 24.05 percentage points, with the clearest gains on Arrange Mouse and Retrieve Drink. Pack Tennis Canister benefits relatively little. We hypothesize that faster motions amplify tracking errors caused by joint response delays. Arrange Mouse requires large, coordinated motions while avoiding the chair, making accurate timing across joints particularly important. For slower, posture-dominated motions, such as reaching into the box in Pack Tennis Canister, these dynamic differences may have less impact on policy performance.

\subsection{REAM Improves Performance with All Camera Views}
The observation ablations in Figure~\ref{fig:observation-results} highlight the importance of head-camera context for whole-body manipulation. Although the wrist-only condition performs well on Pack Tennis Canister and Retrieve Drink, it achieves only 20\% task progress on Arrange Mouse, compared with 82.5\% for EAR only using all cameras. During parts of Arrange Mouse, the arms are positioned in front of the chair, leaving this obstacle outside the wrist-camera views. The head camera provides the broader scene context needed to coordinate motion around it. Consequently, relying only on wrist observations produces uneven performance across tasks with different visual-context requirements.

Retaining all cameras without REAM performs well on Arrange Mouse and Pack Tennis Canister, but underperforms the wrist-only condition on Retrieve Drink. The robot's kinematic constraints prevent accurate reproduction of the demonstrated head poses, creating a mismatch between human training images and robot observations during rollout. In Retrieve Drink, EAR-only failures frequently occur at the task stage corresponding to the largest head-pose errors in the demonstrations. This correspondence is consistent with our hypothesis that viewpoint mismatch contributes to the performance degradation.

To examine this tradeoff, we relax the head-tracking objective to prioritize wrist tracking while retaining all camera observations. Performance degrades further across all three tasks relative to EAR only. These results suggest that prioritizing wrist tracking at the expense of head alignment does not resolve the observation mismatch when the policy continues to receive the head view.

Combining all cameras with REAM achieves the highest average task progress among the observation conditions, at 91.79\%, an improvement of 14.05 percentage points over EAR only. REAM recovers most of the performance lost on Retrieve Drink while retaining the head-camera context needed for Arrange Mouse; performance on Pack Tennis Canister remains comparable. This provides strong performance across tasks without removing a camera from the policy.

Figure~\ref{fig:ream} provides a qualitative example from a Retrieve Drink training demonstration. Relative attention to the head camera decreases during the interval with the largest head-pose error. This aligns with the intended role of REAM: reducing reliance on poorly matched views at particular moments in a trajectory while preserving access to useful context from the available cameras.

\subsection{DAgger Improves Performance Across Rounds}
On Fold Box and Hang Jacket, each round of DAgger improves policy performance, as shown in Figure~\ref{fig:dagger-results}. With Full rollouts, average task progress across the two tasks increases from 6.82\% for the initial human-trained policy to 50.68\% after the first round and 79.77\% after the second. Interventions only also improves substantially, reaching 47.27\% and 69.77\% after the corresponding rounds. Retaining complete corrected rollouts therefore yields higher average performance than training only on intervention segments.

We observe that each round enables the policy to complete additional task stages, allowing subsequent corrections to address failures later in the task. This progression helps explain the approximately linear increase in task progress over the two evaluated rounds: successive rounds extend how far the policy can execute the task successfully. We expect longer tasks with more subtask stages to require additional rounds of correction to achieve comparable progress.

Robot-only demos also performs strongly, achieving 73.41\% average task progress using the same collected robot rollouts without human demonstrations in the final training dataset, compared with 79.77\% when human demonstrations are retained. However, these rollouts were collected by correcting policies initially trained on human demonstrations. Collecting successful robot demonstrations for these tasks is often difficult without this initial policy, so human demonstrations remain valuable for making correction-data collection practical.

\raggedbottom
\section{Related Work}
\label{sec:related}

\subsection{Sensorized Human Demonstrations and Embodiment Transfer}
Sensorized human interfaces enable demonstration collection without requiring a robot at the collection site. Gripper-based interfaces reduce the embodiment gap by aligning the demonstration tool and camera placement with the robot's manipulation interface, with subsequent designs improving tracking, ergonomics, and collection scalability~\cite{song2020grasping,pmlr-v155-young21a,shafiullah2023bringing,chi2024universal,pmlr-v305-zhaxizhuoma25a,ohkawa2026yubi}. Wearable interfaces offer alternative tradeoffs: exoskeletons align recorded motion with robot kinematics~\cite{fang2024airexo,fang2025airexo2,xu2025dexumi}, while instrumented gloves capture natural hand motion for subsequent retargeting~\cite{Wang-RSS-24,TaoT-RSS-25}. Egocentric bare-hand capture further reduces instrumentation, using human--robot co-training to connect human observations and motion to robot behavior~\cite{kareer2025egomimic}. Related industrial efforts also emphasize scalable collection of sensorized-interaction data~\cite{sunday2025act1,generalist2025gen0}.

Beyond hand-centric manipulation, interfaces that jointly record head and hand motion provide broader scene context and demonstrations of active perception~\cite{zeng2025activeumi,yu2025egomi}. Mobile manipulation systems combine human motion with robot data or whole-body control to coordinate navigation and manipulation~\cite{zhu2026emma,xu2026hommi}. Humanoid extensions explore both additional pelvis and foot tracking to specify body motion and simpler head-and-hand interfaces that delegate the remaining coordination to learned controllers~\cite{nai2026humanoid,zhao2026halomi}.

Transferring these demonstrations requires aligning both observations and actions across embodiments. Visual adaptation methods replace human appearance with robot imagery~\cite{fang2025airexo2,xu2025dexumi}, mask embodiment-specific features or use geometry-aware representations~\cite{kareer2025egomimic,xu2026hommi}, and reproject observations toward the robot viewpoint~\cite{zhao2026halomi}. Motion adaptation addresses robot kinematic constraints through retargeting and constrained control~\cite{zhu2026emma,xu2026hommi}, while latency compensation and controller-aware trajectory adjustment account for differences between recorded motion and executed actions~\cite{chi2024universal,zhao2026halomi}. Force-sensing interfaces additionally capture interaction wrenches for learning force-aware and compliant manipulation~\cite{liu2025forcemimic,choi2026umift}.

Our work complements these approaches through scene-aware whole-body retargeting, retargeting-error-guided camera masking, and delay-calibrated action assignment, followed by on-policy correction.

\subsection{Learning for Whole-Body Mobile Manipulation}
Behavior cloning naturally extends to mobile manipulation, provided that suitable demonstration data can be collected. Whole-body teleoperation provides demonstrations of coordinated base and arm motion, enabling learned policies to perform diverse household tasks~\cite{fu2024mobile,wu2024tidybotopensourceholonomicmobile,jiang2025behavior}. Action-chunking and diffusion-based policies provide effective architectures for learning visuomotor behavior from these demonstrations~\cite{zhao2023learning,Chi-RSS-23}. Simulation can further expand source demonstrations across new task configurations while accounting for reachability and camera visibility~\cite{li2026momagen}.

Human demonstrations provide a complementary source of supervision without requiring mobile robot teleoperation. Several approaches learn task-space policies from human demonstrations and use inverse kinematics or whole-body controllers to execute their predictions~\cite{ha2025umilegs,seo2025legato,xu2026hommi}. Other methods combine human mobile-manipulation data with stationary robot demonstrations~\cite{zhu2026emma}, or extend human demonstration learning to humanoids through hierarchical policies and learned motion controllers~\cite{nai2026humanoid,zhao2026halomi}. Offline retargeting instead converts recorded human motion into robot joint trajectories before policy training~\cite{chen2026warp}.

Reinforcement learning provides another route to acquiring coordinated mobile-manipulation behavior. Simulation-trained policies learn base placement and navigation conditioned on manipulation goals~\cite{honerkamp2023n2m2,jauhri2022reachability}, coupled locomotion and arm control~\cite{fu2023deepwholebody}, and visual coordination of movement, manipulation, and perception~\cite{liu2025visualwholebody,uppal2024spin}. Hybrid approaches combine a demonstration-trained visual planner with an RL controller~\cite{he2024learningvisualquadrupedallocomanipulation}. COWBOY combines environment-aware whole-body control for reaching with local RL experts for object interaction, then distills their behavior into a single visual whole-body policy~\cite{yang2026cowboy}. Beyond simulation, autonomous real-world RL has also enabled navigation and grasping to improve through robot interaction~\cite{sun2022relmm}. 

Our work focuses on learning whole-body policies from sensorized human demonstrations, using environment-aware retargeting to construct robot motion supervision and on-policy correction to improve performance.

\subsection{Continual Improvement of Robot Policies}
Once an initial policy is available, further interaction can improve its reliability and task performance. Some reinforcement learning approaches learn residual corrections around a frozen behavior-cloning policy~\cite{ankile2024resip,ankile2025resfit}, or combine learned action edits with value-based action selection~\cite{dong2026expo,dong2026expoft}. Other approaches directly fine-tune diffusion policies~\cite{ren2025dppo,lei2025rl100} or train lightweight actor--critic heads using pretrained VLA representations~\cite{xu2026rltoken}. RL can also generate data for subsequent supervised refinement, using residual specialists to collect recovery trajectories that are distilled back into a generalist policy~\cite{xiao2026pld}.

Interactive imitation learning improves policies through expert feedback on their own execution. DAgger addresses distribution shift by collecting expert action labels at states visited by the learned policy and iteratively training on the aggregated dataset~\cite{ross2011dagger}. Subsequent methods make this process more practical by selectively requesting supervision. Human-gated approaches let operators decide when to take control~\cite{kelly2019hgdagger}, while robot-gated approaches use novelty and estimated risk to request interventions within a supervision budget~\cite{hoque2022thriftydagger}.

For manipulation, the correction interface determines what feedback humans can provide. Compliant bilateral teleoperation supports smooth handovers between autonomous execution and human control~\cite{wu2025robocopilot}, while kinesthetic interfaces allow humans to provide residual motion and force corrections during ongoing policy execution~\cite{xu2025crdagger}. Correction protocols also shape the behaviors learned: combining recovery to familiar states with corrective task completion teaches policies to retry and adapt during long-horizon manipulation~\cite{hu2025rac}.

Another question is how to learn from rollouts containing both autonomous rollouts and human corrections. Weighted imitation methods balance these data sources and use intervention signals to prioritize informative behavior~\cite{mandlekar2020iwr,liu2023sirius}. Preference-based approaches additionally use these signals to reinforce corrective actions and suppress failure-prone behavior~\cite{xia2025apo}. Distributed systems extend this improvement loop across robot fleets, coupling experience collection with frequent policy updates through either interactive imitation learning or RL~\cite{pan2026sop}. We study interactive improvement for whole-body policies initialized from sensorized human demonstrations, comparing aggregation of complete corrected rollouts with intervention segments alone.

\section{Conclusion and Limitations}
\label{sec:conclusion}
\label{sec:limitations}
We present Pixel-UMI, a pipeline for learning whole-body mobile manipulation from sensorized human demonstrations. Our results highlight the value of absolute joint-space actions for preserving body configurations that head and end-effector targets leave unspecified, and of environment geometry for determining those configurations. Together with calibrated action assignment and REAM, this approach supports challenging mobile and dexterous manipulation, achieving 87.0\% average task progress across five tasks. For contact-rich tasks, on-policy corrections progressively extend the stages the policy can complete, increasing average task progress from 6.82\% to 79.77\% over two rounds. These findings demonstrate how human demonstrations can provide useful whole-body motion supervision and a starting point for further improvement through robot interaction.

Several limitations remain. Delay-calibrated action assignment provides approximate command labels but cannot recover the contact-dependent actions underlying demonstrated motion. In particular, motion alone does not reveal the forces applied when contact prevents further movement. Force/torque or tactile sensing, combined with robot dynamics and controller models, could improve action inference and reduce reliance on corrective robot data.

Retargeting quality also depends on the reconstructed environment. Our reconstruction uses a single head-mounted depth camera, leaving occluded or unobserved surfaces incomplete; additional viewpoints and sensors could improve coverage and accuracy. The static-scene formulation further limits how moving obstacles and intended contacts are represented. Finally, although retargeting generates whole-body trajectories, the demonstrated motion targets specify only the head and grippers. Intentional contacts involving other body parts, such as using an elbow to hold a door open or a hip to close a drawer, are therefore not explicitly captured. Supporting these behaviors would require richer body-motion or contact supervision and retargeting objectives that account for the intended interactions.

\section*{Acknowledgment}
We would like to thank Philip Han and Ray Liu for their help with data collection. We would also like to thank Kenneth Shaw, Arthur Allshire, Ritvik Singh, Andrew Wang, Zheyuan Hu, Peiqi Liu, Sri Anumakonda, Steven Oh, Adam Kan, Haoran Huan, Jiahui Yang, Hengkai Pan, Lili Chen, Mihir Prabhudesai, Wenli Xiao, and Tairan He for their thoughtful discussions. This work was supported by AFOSR FA9550-23-1-0747, ONR MURI N00014-22-1-2773 and ONR MURI N00014-24-1-2748.
\begingroup
\renewcommand{\baselinestretch}{0.98}
\bibliographystyle{IEEEtran}
\bibliography{references}

\begin{thebibliography}{10}
\providecommand{\url}[1]{#1}
\csname url@samestyle\endcsname
\providecommand{\newblock}{\relax}
\providecommand{\bibinfo}[2]{#2}
\providecommand{\BIBentrySTDinterwordspacing}{\spaceskip=0pt\relax}
\providecommand{\BIBentryALTinterwordstretchfactor}{4}
\providecommand{\BIBentryALTinterwordspacing}{\spaceskip=\fontdimen2\font plus
\BIBentryALTinterwordstretchfactor\fontdimen3\font minus \fontdimen4\font\relax}
\providecommand{\BIBforeignlanguage}[2]{{%
\expandafter\ifx\csname l@#1\endcsname\relax
\typeout{** WARNING: IEEEtran.bst: No hyphenation pattern has been}%
\typeout{** loaded for the language `#1'. Using the pattern for}%
\typeout{** the default language instead.}%
\else
\language=\csname l@#1\endcsname
\fi
#2}}
\providecommand{\BIBdecl}{\relax}
\BIBdecl

\bibitem{fu2024mobile}
\BIBentryALTinterwordspacing
Z.~Fu, T.~Z. Zhao, and C.~Finn, ``Mobile {ALOHA}: Learning bimanual mobile manipulation with low-cost whole-body teleoperation,'' in \emph{Conference on Robot Learning}, 2024. [Online]. Available: \url{https://mobile-aloha.github.io/}
\BIBentrySTDinterwordspacing

\bibitem{jiang2025behavior}
\BIBentryALTinterwordspacing
Y.~Jiang, R.~Zhang, J.~Wong, C.~Wang, Y.~Ze, H.~Yin, C.~Gokmen, S.~Song, J.~Wu, and L.~Fei-Fei, ``{BEHAVIOR} robot suite: Streamlining real-world whole-body manipulation for everyday household activities,'' in \emph{Proceedings of The 9th Conference on Robot Learning}, ser. Proceedings of Machine Learning Research, vol. 305, 2025, pp. 1246--1281. [Online]. Available: \url{https://proceedings.mlr.press/v305/jiang25b.html}
\BIBentrySTDinterwordspacing

\bibitem{chi2024universal}
\BIBentryALTinterwordspacing
C.~Chi, Z.~Xu, C.~Pan, E.~Cousineau, B.~Burchfiel, S.~Feng, R.~Tedrake, and S.~Song, ``Universal manipulation interface: In-the-wild robot teaching without in-the-wild robots,'' in \emph{Robotics: Science and Systems}, 2024. [Online]. Available: \url{https://umi-gripper.github.io/}
\BIBentrySTDinterwordspacing

\bibitem{sunday2025act1}
\BIBentryALTinterwordspacing
{Sunday Team}, ``{ACT-1}: A robot foundation model trained on zero robot data,'' Sunday Robotics research blog, Nov. 2025. [Online]. Available: \url{https://www.sunday.ai/blog/no-robot-data}
\BIBentrySTDinterwordspacing

\bibitem{xu2026hommi}
\BIBentryALTinterwordspacing
X.~Xu, J.~Park, H.~Zhang, E.~Cousineau, A.~Bhat, J.~Barreiros, D.~Wang, J.~Bohg, and S.~Song, ``{HoMMI}: Learning whole-body mobile manipulation from human demonstrations,'' \emph{arXiv preprint arXiv:2603.03243}, 2026. [Online]. Available: \url{https://arxiv.org/abs/2603.03243}
\BIBentrySTDinterwordspacing

\bibitem{zeng2025activeumi}
\BIBentryALTinterwordspacing
Q.~Zeng, C.~Li, J.~St.~John, Z.~Zhou, J.~Wen, G.~Feng, Y.~Zhu, and Y.~Xu, ``{ActiveUMI}: Robotic manipulation with active perception from robot-free human demonstrations,'' \emph{arXiv preprint arXiv:2510.01607}, 2025. [Online]. Available: \url{https://arxiv.org/abs/2510.01607}
\BIBentrySTDinterwordspacing

\bibitem{xie2020fabrics}
\BIBentryALTinterwordspacing
M.~Xie, K.~Van~Wyk, A.~Li, M.~A. Rana, Q.~Wan, D.~Fox, B.~Boots, and N.~Ratliff, ``Geometric fabrics for the acceleration-based design of robotic motion,'' \emph{arXiv preprint arXiv:2010.14750}, 2020. [Online]. Available: \url{https://arxiv.org/abs/2010.14750}
\BIBentrySTDinterwordspacing

\bibitem{liu2025forcemimic}
\BIBentryALTinterwordspacing
W.~Liu, J.~Wang, Y.~Wang, W.~Wang, and C.~Lu, ``{ForceMimic}: Force-centric imitation learning with force-motion capture system for contact-rich manipulation,'' in \emph{IEEE International Conference on Robotics and Automation}, 2025. [Online]. Available: \url{https://arxiv.org/abs/2410.07554}
\BIBentrySTDinterwordspacing

\bibitem{choi2026umift}
\BIBentryALTinterwordspacing
H.~Choi, Y.~Hou, C.~Pan, S.~Hong, A.~Patel, X.~Xu, M.~R. Cutkosky, and S.~Song, ``In-the-wild compliant manipulation with {UMI-FT},'' \emph{arXiv preprint arXiv:2601.09988}, 2026. [Online]. Available: \url{https://arxiv.org/abs/2601.09988}
\BIBentrySTDinterwordspacing

\bibitem{ross2011dagger}
\BIBentryALTinterwordspacing
S.~Ross, G.~Gordon, and D.~Bagnell, ``A reduction of imitation learning and structured prediction to no-regret online learning,'' in \emph{Proceedings of the Fourteenth International Conference on Artificial Intelligence and Statistics}, ser. Proceedings of Machine Learning Research, vol.~15, 2011, pp. 627--635. [Online]. Available: \url{https://proceedings.mlr.press/v15/ross11a.html}
\BIBentrySTDinterwordspacing

\bibitem{pixel2027system}
\BIBentryALTinterwordspacing
T.~Tao, J.~J. Liu, R.~Salakhutdinov, and D.~Pathak, ``{Pixel}: A co-designed full-stack robot system for learning general-purpose mobile manipulation,'' 2026, preprint. [Online]. Available: \url{https://pixel-robot.com/pixel/}
\BIBentrySTDinterwordspacing

\bibitem{besl1992registration}
P.~J. Besl and N.~D. McKay, ``A method for registration of {3-D} shapes,'' \emph{IEEE Transactions on Pattern Analysis and Machine Intelligence}, vol.~14, no.~2, pp. 239--256, 1992.

\bibitem{chen1992object}
Y.~Chen and G.~Medioni, ``Object modelling by registration of multiple range images,'' \emph{Image and Vision Computing}, vol.~10, no.~3, pp. 145--155, 1992.

\bibitem{choi2015robust}
\BIBentryALTinterwordspacing
S.~Choi, Q.-Y. Zhou, and V.~Koltun, ``Robust reconstruction of indoor scenes,'' in \emph{Proceedings of the IEEE Conference on Computer Vision and Pattern Recognition}, 2015, pp. 5556--5565. [Online]. Available: \url{https://openaccess.thecvf.com/content_cvpr_2015/html/Choi_Robust_Reconstruction_of_2015_CVPR_paper.html}
\BIBentrySTDinterwordspacing

\bibitem{hornung2013octomap}
\BIBentryALTinterwordspacing
A.~Hornung, K.~M. Wurm, M.~Bennewitz, C.~Stachniss, and W.~Burgard, ``{OctoMap}: An efficient probabilistic {3D} mapping framework based on octrees,'' \emph{Autonomous Robots}, 2013. [Online]. Available: \url{https://octomap.github.io/}
\BIBentrySTDinterwordspacing

\bibitem{schauer2018peopleremover}
\BIBentryALTinterwordspacing
J.~Schauer and A.~N{\"u}chter, ``The peopleremover---removing dynamic objects from {3-D} point cloud data by traversing a voxel occupancy grid,'' \emph{IEEE Robotics and Automation Letters}, 2018. [Online]. Available: \url{https://robotik.informatik.uni-wuerzburg.de/telematics/download/RA-L_2018/schauer-ral18-peopleremover.pdf}
\BIBentrySTDinterwordspacing

\bibitem{zakka2026mink}
\BIBentryALTinterwordspacing
K.~Zakka, ``{Mink}: Python inverse kinematics based on {MuJoCo},'' Software, version 1.1.0, 2026. [Online]. Available: \url{https://github.com/kevinzakka/mink}
\BIBentrySTDinterwordspacing

\bibitem{caron2026pink}
\BIBentryALTinterwordspacing
S.~Caron, Y.~{De Mont-Marin}, R.~Budhiraja, S.~H. Bang, I.~Domrachev, S.~Nedelchev, P.~Du, J.~Vaillant, B.~Wingo, S.~Patapati, D.~{San Jos{\'e} Pro}, and N.~G. {Marticorena Vidal}, ``{Pink}: Python inverse kinematics based on {Pinocchio},'' Software, version 4.4.0, 2026. [Online]. Available: \url{https://github.com/pink-kinematics/pink}
\BIBentrySTDinterwordspacing

\bibitem{sunday2026technology}
\BIBentryALTinterwordspacing
{Sunday Robotics}, ``Technology,'' Technical overview, accessed October 19, 2026. [Online]. Available: \url{https://www.sunday.ai/technology}
\BIBentrySTDinterwordspacing

\bibitem{kelly2019hgdagger}
\BIBentryALTinterwordspacing
M.~Kelly, C.~Sidrane, K.~Driggs-Campbell, and M.~J. Kochenderfer, ``{HG-DAgger}: Interactive imitation learning with human experts,'' \emph{arXiv preprint arXiv:1810.02890}, 2019. [Online]. Available: \url{https://arxiv.org/abs/1810.02890}
\BIBentrySTDinterwordspacing

\bibitem{hu2025rac}
\BIBentryALTinterwordspacing
Z.~Hu, R.~Wu, N.~Enock, J.~Li, R.~Kadakia, Z.~Erickson, and A.~Kumar, ``{RaC}: Robot learning for long-horizon tasks by scaling recovery and correction,'' \emph{arXiv preprint arXiv:2509.07953}, 2025. [Online]. Available: \url{https://arxiv.org/abs/2509.07953}
\BIBentrySTDinterwordspacing

\bibitem{black2025training}
\BIBentryALTinterwordspacing
K.~Black, A.~Z. Ren, M.~Equi, and S.~Levine, ``Training-time action conditioning for efficient real-time chunking,'' \emph{arXiv preprint arXiv:2512.05964}, 2025. [Online]. Available: \url{https://arxiv.org/abs/2512.05964}
\BIBentrySTDinterwordspacing

\bibitem{simeoni2025dinov3}
\BIBentryALTinterwordspacing
O.~Sim{\'e}oni, H.~V. Vo, M.~Seitzer, F.~Baldassarre, M.~Oquab, C.~Jose, V.~Khalidov, M.~Szafraniec, S.~Yi, M.~Ramamonjisoa, F.~Massa, D.~Haziza, L.~Wehrstedt, J.~Wang, T.~Darcet, T.~Moutakanni, L.~Sentana, C.~Roberts, A.~Vedaldi, J.~Tolan, J.~Brandt, C.~Couprie, J.~Mairal, H.~J{\'e}gou, P.~Labatut, and P.~Bojanowski, ``{DINOv3},'' \emph{arXiv preprint arXiv:2508.10104}, 2025. [Online]. Available: \url{https://arxiv.org/abs/2508.10104}
\BIBentrySTDinterwordspacing

\bibitem{lipman2023flow}
\BIBentryALTinterwordspacing
Y.~Lipman, R.~T.~Q. Chen, H.~Ben-Hamu, M.~Nickel, and M.~Le, ``Flow matching for generative modeling,'' \emph{arXiv preprint arXiv:2210.02747}, 2023. [Online]. Available: \url{https://arxiv.org/abs/2210.02747}
\BIBentrySTDinterwordspacing

\bibitem{song2020grasping}
\BIBentryALTinterwordspacing
S.~Song, A.~Zeng, J.~Lee, and T.~Funkhouser, ``Grasping in the wild: Learning 6dof closed-loop grasping from low-cost demonstrations,'' \emph{IEEE Robotics and Automation Letters}, vol.~5, no.~3, pp. 4978--4985, 2020. [Online]. Available: \url{https://arxiv.org/abs/1912.04344}
\BIBentrySTDinterwordspacing

\bibitem{pmlr-v155-young21a}
\BIBentryALTinterwordspacing
S.~Young, D.~Gandhi, S.~Tulsiani, A.~Gupta, P.~Abbeel, and L.~Pinto, ``Visual imitation made easy,'' in \emph{Proceedings of the 2020 Conference on Robot Learning}, ser. Proceedings of Machine Learning Research, vol. 155.\hskip 1em plus 0.5em minus 0.4em\relax PMLR, 2021, pp. 1992--2005. [Online]. Available: \url{https://proceedings.mlr.press/v155/young21a.html}
\BIBentrySTDinterwordspacing

\bibitem{shafiullah2023bringing}
\BIBentryALTinterwordspacing
N.~M.~M. Shafiullah, A.~Rai, H.~Etukuru, Y.~Liu, I.~Misra, S.~Chintala, and L.~Pinto, ``On bringing robots home,'' \emph{arXiv preprint arXiv:2311.16098}, 2023. [Online]. Available: \url{https://arxiv.org/abs/2311.16098}
\BIBentrySTDinterwordspacing

\bibitem{pmlr-v305-zhaxizhuoma25a}
\BIBentryALTinterwordspacing
Z.~Zhaxizhuoma, K.~Liu, C.~Guan, Z.~Jia, Z.~Wu, X.~Liu, T.~Wang, S.~Liang, P.~Chen, P.~Zhang, H.~Song, D.~Qu, D.~Wang, Z.~Wang, N.~Cao, Y.~Ding, B.~Zhao, and X.~Li, ``{FastUMI}: A scalable and hardware-independent universal manipulation interface with dataset,'' in \emph{Proceedings of the 9th Conference on Robot Learning}, ser. Proceedings of Machine Learning Research, vol. 305.\hskip 1em plus 0.5em minus 0.4em\relax PMLR, 2025, pp. 3069--3093. [Online]. Available: \url{https://proceedings.mlr.press/v305/zhaxizhuoma25a.html}
\BIBentrySTDinterwordspacing

\bibitem{ohkawa2026yubi}
\BIBentryALTinterwordspacing
T.~Ohkawa, J.~Arima, Y.~Noguchi, M.~Tateno, M.~Sugiura, T.~Okubo, K.~Ikeuchi, Y.~Shin, H.~Nishizawa, N.~Kanazawa, Y.~Wakayama, D.~Fukunaga, K.~Makihara, T.~Motoda, F.~Erich, Y.~Domae, T.~Matsushima, Y.~Okumatsu, and K.~Ota, ``{YUBI}: Yielding universal bidigital interface for bimanual dexterous manipulation at scale,'' 2026. [Online]. Available: \url{https://arxiv.org/abs/2606.10244}
\BIBentrySTDinterwordspacing

\bibitem{fang2024airexo}
\BIBentryALTinterwordspacing
H.~Fang, H.-S. Fang, Y.~Wang, J.~Ren, J.~Chen, R.~Zhang, W.~Wang, and C.~Lu, ``{AirExo}: Low-cost exoskeletons for learning whole-arm manipulation in the wild,'' in \emph{2024 IEEE International Conference on Robotics and Automation (ICRA)}, 2024, pp. 15\,031--15\,038. [Online]. Available: \url{https://doi.org/10.1109/ICRA57147.2024.10610799}
\BIBentrySTDinterwordspacing

\bibitem{fang2025airexo2}
\BIBentryALTinterwordspacing
H.~Fang, C.~Wang, Y.~Wang, J.~Chen, S.~Xia, J.~Lv, Z.~He, X.~Yi, Y.~Guo, X.~Zhan, L.~Yang, W.~Wang, C.~Lu, and H.-S. Fang, ``{AirExo-2}: Scaling up generalizable robotic imitation learning with low-cost exoskeletons,'' \emph{arXiv preprint arXiv:2503.03081}, 2025. [Online]. Available: \url{https://arxiv.org/abs/2503.03081}
\BIBentrySTDinterwordspacing

\bibitem{xu2025dexumi}
\BIBentryALTinterwordspacing
M.~Xu, H.~Zhang, Y.~Hou, Z.~Xu, L.~Fan, M.~Veloso, and S.~Song, ``{DexUMI}: Using human hand as the universal manipulation interface for dexterous manipulation,'' \emph{arXiv preprint arXiv:2505.21864}, 2025. [Online]. Available: \url{https://arxiv.org/abs/2505.21864}
\BIBentrySTDinterwordspacing

\bibitem{Wang-RSS-24}
\BIBentryALTinterwordspacing
C.~Wang, H.~Shi, W.~Wang, R.~Zhang, L.~Fei-Fei, and K.~Liu, ``{DexCap}: Scalable and portable mocap data collection system for dexterous manipulation,'' in \emph{Proceedings of Robotics: Science and Systems}, Delft, Netherlands, Jul. 2024. [Online]. Available: \url{https://www.roboticsproceedings.org/rss20/p043.html}
\BIBentrySTDinterwordspacing

\bibitem{TaoT-RSS-25}
\BIBentryALTinterwordspacing
T.~Tao, M.~K. Srirama, J.~J. Liu, K.~Shaw, and D.~Pathak, ``{DexWild}: Dexterous human interactions for in-the-wild robot policies,'' in \emph{Proceedings of Robotics: Science and Systems}, Los Angeles, CA, USA, Jun. 2025. [Online]. Available: \url{https://www.roboticsproceedings.org/rss21/p075.html}
\BIBentrySTDinterwordspacing

\bibitem{kareer2025egomimic}
\BIBentryALTinterwordspacing
S.~Kareer, D.~Patel, R.~Punamiya, P.~Mathur, S.~Cheng, C.~Wang, J.~Hoffman, and D.~Xu, ``{EgoMimic}: Scaling imitation learning via egocentric video,'' in \emph{2025 IEEE International Conference on Robotics and Automation (ICRA)}, 2025, pp. 13\,226--13\,233. [Online]. Available: \url{https://arxiv.org/abs/2410.24221}
\BIBentrySTDinterwordspacing

\bibitem{generalist2025gen0}
\BIBentryALTinterwordspacing
{Generalist Team}, ``{GEN-0}: Embodied foundation models that scale with physical interaction,'' \emph{Generalist AI Blog}, Nov. 2025, first-party technical blog; not peer reviewed. [Online]. Available: \url{https://generalistai.com/blog/gen-0}
\BIBentrySTDinterwordspacing

\bibitem{yu2025egomi}
\BIBentryALTinterwordspacing
J.~Yu, Y.~Shentu, D.~Wu, P.~Abbeel, K.~Goldberg, and P.~Wu, ``{EgoMI}: Learning active vision and whole-body manipulation from egocentric human demonstrations,'' 2025. [Online]. Available: \url{https://arxiv.org/abs/2511.00153}
\BIBentrySTDinterwordspacing

\bibitem{zhu2026emma}
\BIBentryALTinterwordspacing
L.~Y. Zhu, P.~Kuppili, R.~Punamiya, P.~Aphiwetsa, D.~Patel, S.~Kareer, S.~Ha, and D.~Xu, ``{EMMA}: Scaling mobile manipulation via egocentric human data,'' \emph{IEEE Robotics and Automation Letters}, vol.~11, no.~3, pp. 3087--3094, 2026. [Online]. Available: \url{https://doi.org/10.1109/LRA.2026.3653320}
\BIBentrySTDinterwordspacing

\bibitem{nai2026humanoid}
\BIBentryALTinterwordspacing
R.~Nai, B.~Zheng, J.~Zhao, H.~Zhu, S.~Dai, Z.~Chen, Y.~Hu, Y.~Hu, T.~Zhang, C.~Wen, and Y.~Gao, ``Humanoid manipulation interface: Humanoid whole-body manipulation from robot-free demonstrations,'' \emph{arXiv preprint arXiv:2602.06643}, 2026. [Online]. Available: \url{https://arxiv.org/abs/2602.06643}
\BIBentrySTDinterwordspacing

\bibitem{zhao2026halomi}
\BIBentryALTinterwordspacing
Z.~Zhao, Y.~Zhao, G.~Zhang, C.~Liu, M.~Zheng, and W.~Lian, ``{HALOMI}: Learning humanoid loco-manipulation with active perception from human demonstrations,'' 2026. [Online]. Available: \url{https://arxiv.org/abs/2606.18772}
\BIBentrySTDinterwordspacing

\bibitem{wu2024tidybotopensourceholonomicmobile}
\BIBentryALTinterwordspacing
J.~Wu, W.~Chong, R.~Holmberg, A.~Prasad, Y.~Gao, O.~Khatib, S.~Song, S.~Rusinkiewicz, and J.~Bohg, ``{TidyBot++}: An open-source holonomic mobile manipulator for robot learning,'' in \emph{Proceedings of the 8th Conference on Robot Learning}, ser. Proceedings of Machine Learning Research, vol. 270.\hskip 1em plus 0.5em minus 0.4em\relax PMLR, 2025, pp. 3729--3741. [Online]. Available: \url{https://proceedings.mlr.press/v270/wu25a.html}
\BIBentrySTDinterwordspacing

\bibitem{zhao2023learning}
\BIBentryALTinterwordspacing
T.~Z. Zhao, V.~Kumar, S.~Levine, and C.~Finn, ``Learning fine-grained bimanual manipulation with low-cost hardware,'' in \emph{Proceedings of Robotics: Science and Systems}, 2023. [Online]. Available: \url{https://roboticsproceedings.org/rss19/p016.html}
\BIBentrySTDinterwordspacing

\bibitem{Chi-RSS-23}
C.~Chi, S.~Feng, Y.~Du, Z.~Xu, E.~Cousineau, B.~C. Burchfiel, and S.~Song, ``{Diffusion Policy: Visuomotor Policy Learning via Action Diffusion},'' in \emph{Proceedings of Robotics: Science and Systems}, Daegu, Republic of Korea, Jul. 2023.

\bibitem{li2026momagen}
\BIBentryALTinterwordspacing
C.~Li, M.~Xu, A.~Bahety, H.~Yin, Y.~Jiang, H.~Huang, J.~Wong, S.~Garlanka, C.~Gokmen, R.~Zhang, W.~Liu, J.~Wu, R.~Mart{\'i}n-Mart{\'i}n, and L.~Fei-Fei, ``{MoMaGen}: Generating demonstrations under soft and hard constraints for multi-step bimanual mobile manipulation,'' in \emph{International Conference on Learning Representations}, 2026. [Online]. Available: \url{https://arxiv.org/abs/2510.18316}
\BIBentrySTDinterwordspacing

\bibitem{ha2025umilegs}
\BIBentryALTinterwordspacing
H.~Ha, Y.~Gao, Z.~Fu, J.~Tan, and S.~Song, ``{UMI-on-Legs}: Making manipulation policies mobile with manipulation-centric whole-body controllers,'' in \emph{Proceedings of The 8th Conference on Robot Learning}, ser. Proceedings of Machine Learning Research, vol. 270, 2025, pp. 5254--5270. [Online]. Available: \url{https://proceedings.mlr.press/v270/ha25a.html}
\BIBentrySTDinterwordspacing

\bibitem{seo2025legato}
\BIBentryALTinterwordspacing
M.~Seo, H.~A. Park, S.~Yuan, Y.~Zhu, and L.~Sentis, ``{LEGATO}: Cross-embodiment imitation using a grasping tool,'' \emph{IEEE Robotics and Automation Letters}, vol.~10, no.~3, pp. 2854--2861, 2025. [Online]. Available: \url{https://arxiv.org/abs/2411.03682}
\BIBentrySTDinterwordspacing

\bibitem{chen2026warp}
\BIBentryALTinterwordspacing
Z.~Chen, C.~Kong, C.~Zhang, Y.~Yang, L.~Y. Zhu, S.~Kousik, and D.~Xu, ``{WARP}: Whole-body retargeting for learning from offline human demonstrations,'' \emph{arXiv preprint arXiv:2606.29940}, 2026. [Online]. Available: \url{https://arxiv.org/abs/2606.29940}
\BIBentrySTDinterwordspacing

\bibitem{honerkamp2023n2m2}
\BIBentryALTinterwordspacing
D.~Honerkamp, T.~Welschehold, and A.~Valada, ``{N$^2$M$^2$}: Learning navigation for arbitrary mobile manipulation motions in unseen and dynamic environments,'' \emph{IEEE Transactions on Robotics}, vol.~39, no.~5, pp. 3601--3619, 2023. [Online]. Available: \url{https://arxiv.org/abs/2206.08737}
\BIBentrySTDinterwordspacing

\bibitem{jauhri2022reachability}
\BIBentryALTinterwordspacing
S.~Jauhri, J.~Peters, and G.~Chalvatzaki, ``Robot learning of mobile manipulation with reachability behavior priors,'' \emph{IEEE Robotics and Automation Letters}, vol.~7, no.~3, pp. 8399--8406, 2022. [Online]. Available: \url{https://arxiv.org/abs/2203.04051}
\BIBentrySTDinterwordspacing

\bibitem{fu2023deepwholebody}
\BIBentryALTinterwordspacing
Z.~Fu, X.~Cheng, and D.~Pathak, ``Deep whole-body control: Learning a unified policy for manipulation and locomotion,'' in \emph{Proceedings of The 6th Conference on Robot Learning}, ser. Proceedings of Machine Learning Research, vol. 205, 2023, pp. 138--149. [Online]. Available: \url{https://proceedings.mlr.press/v205/fu23a.html}
\BIBentrySTDinterwordspacing

\bibitem{liu2025visualwholebody}
\BIBentryALTinterwordspacing
M.~Liu, Z.~Chen, X.~Cheng, Y.~Ji, R.-Z. Qiu, R.~Yang, and X.~Wang, ``Visual whole-body control for legged loco-manipulation,'' in \emph{Proceedings of The 8th Conference on Robot Learning}, ser. Proceedings of Machine Learning Research, vol. 270, 2025, pp. 234--257. [Online]. Available: \url{https://proceedings.mlr.press/v270/liu25b.html}
\BIBentrySTDinterwordspacing

\bibitem{uppal2024spin}
\BIBentryALTinterwordspacing
S.~Uppal, A.~Agarwal, H.~Xiong, K.~Shaw, and D.~Pathak, ``{SPIN}: Simultaneous perception, interaction and navigation,'' in \emph{Proceedings of the IEEE/CVF Conference on Computer Vision and Pattern Recognition}, 2024, pp. 18\,133--18\,142. [Online]. Available: \url{https://arxiv.org/abs/2405.07991}
\BIBentrySTDinterwordspacing

\bibitem{he2024learningvisualquadrupedallocomanipulation}
\BIBentryALTinterwordspacing
Z.~He, K.~Lei, Y.~Ze, K.~Sreenath, Z.~Li, and H.~Xu, ``Learning visual quadrupedal loco-manipulation from demonstrations,'' in \emph{2024 IEEE/RSJ International Conference on Intelligent Robots and Systems (IROS)}, 2024, pp. 9102--9109. [Online]. Available: \url{https://arxiv.org/abs/2403.20328}
\BIBentrySTDinterwordspacing

\bibitem{yang2026cowboy}
\BIBentryALTinterwordspacing
J.~Yang, J.~J. Liu, R.~M. Liu, A.~E. Wang, K.~Shaw, D.~Pathak, and R.~Salakhutdinov, ``{COWBOY}: A scalable sim-to-real framework for learning contextual whole-body manipulation,'' in \emph{Conference on Robot Learning}, 2026. [Online]. Available: \url{https://arxiv.org/abs/2610.28948}
\BIBentrySTDinterwordspacing

\bibitem{sun2022relmm}
\BIBentryALTinterwordspacing
C.~Sun, J.~Orbik, C.~M. Devin, B.~H. Yang, A.~Gupta, G.~Berseth, and S.~Levine, ``Fully autonomous real-world reinforcement learning with applications to mobile manipulation,'' in \emph{Proceedings of the 5th Conference on Robot Learning}, ser. Proceedings of Machine Learning Research, vol. 164, 2022, pp. 308--319. [Online]. Available: \url{https://proceedings.mlr.press/v164/sun22a.html}
\BIBentrySTDinterwordspacing

\bibitem{ankile2024resip}
\BIBentryALTinterwordspacing
L.~Ankile, A.~Simeonov, I.~Shenfeld, M.~Torne, and P.~Agrawal, ``From imitation to refinement---residual {RL} for precise assembly,'' \emph{arXiv preprint arXiv:2407.16677}, 2024. [Online]. Available: \url{https://arxiv.org/abs/2407.16677}
\BIBentrySTDinterwordspacing

\bibitem{ankile2025resfit}
\BIBentryALTinterwordspacing
L.~Ankile, Z.~Jiang, R.~Duan, G.~Shi, P.~Abbeel, and A.~Nagabandi, ``Residual off-policy {RL} for finetuning behavior cloning policies,'' \emph{arXiv preprint arXiv:2509.19301}, 2025. [Online]. Available: \url{https://arxiv.org/abs/2509.19301}
\BIBentrySTDinterwordspacing

\bibitem{dong2026expo}
\BIBentryALTinterwordspacing
P.~Dong, Q.~Li, D.~Sadigh, and C.~Finn, ``{EXPO}: Stable reinforcement learning with expressive policies,'' in \emph{International Conference on Learning Representations}, 2026. [Online]. Available: \url{https://arxiv.org/abs/2507.07986}
\BIBentrySTDinterwordspacing

\bibitem{dong2026expoft}
\BIBentryALTinterwordspacing
P.~Dong, K.-H. Hung, T.~Gao, D.~Sadigh, and C.~Finn, ``{EXPO-FT}: Sample-efficient reinforcement learning finetuning for vision-language-action models,'' \emph{arXiv preprint arXiv:2605.25477}, 2026. [Online]. Available: \url{https://arxiv.org/abs/2605.25477}
\BIBentrySTDinterwordspacing

\bibitem{ren2025dppo}
\BIBentryALTinterwordspacing
A.~Z. Ren, J.~Lidard, L.~L. Ankile, A.~Simeonov, P.~Agrawal, A.~Majumdar, B.~Burchfiel, H.~Dai, and M.~Simchowitz, ``Diffusion policy policy optimization,'' in \emph{International Conference on Learning Representations}, 2025. [Online]. Available: \url{https://arxiv.org/abs/2409.00588}
\BIBentrySTDinterwordspacing

\bibitem{lei2025rl100}
\BIBentryALTinterwordspacing
K.~Lei, H.~Li, D.~Yu, Z.~Wei, L.~Guo, Z.~Jiang, Z.~Wang, S.~Liang, and H.~Xu, ``{RL-100}: Performant robotic manipulation with real-world reinforcement learning,'' \emph{arXiv preprint arXiv:2510.14830}, 2025. [Online]. Available: \url{https://arxiv.org/abs/2510.14830}
\BIBentrySTDinterwordspacing

\bibitem{xu2026rltoken}
\BIBentryALTinterwordspacing
C.~Xu, J.~T. Springenberg, M.~Equi, A.~Amin, A.~Esmail, S.~Levine, and L.~Ke, ``{RL Token}: Bootstrapping online {RL} with vision-language-action models,'' \emph{arXiv preprint arXiv:2604.23073}, 2026. [Online]. Available: \url{https://arxiv.org/abs/2604.23073}
\BIBentrySTDinterwordspacing

\bibitem{xiao2026pld}
\BIBentryALTinterwordspacing
W.~Xiao, H.~Lin, A.~Peng, H.~Xue, T.~He, Y.~Xie, F.~Hu, J.~Wu, Z.~Luo, L.~J. Fan, G.~Shi, and Y.~Zhu, ``Self-improving vision-language-action models with data generation via residual {RL},'' in \emph{International Conference on Learning Representations}, 2026. [Online]. Available: \url{https://arxiv.org/abs/2511.00091}
\BIBentrySTDinterwordspacing

\bibitem{hoque2022thriftydagger}
\BIBentryALTinterwordspacing
R.~Hoque, A.~Balakrishna, E.~Novoseller, A.~Wilcox, D.~S. Brown, and K.~Goldberg, ``{ThriftyDAgger}: Budget-aware novelty and risk gating for interactive imitation learning,'' in \emph{Proceedings of the 5th Conference on Robot Learning}, ser. Proceedings of Machine Learning Research, vol. 164, 2022, pp. 598--608. [Online]. Available: \url{https://proceedings.mlr.press/v164/hoque22a.html}
\BIBentrySTDinterwordspacing

\bibitem{wu2025robocopilot}
\BIBentryALTinterwordspacing
P.~Wu, Y.~Shentu, Q.~Liao, D.~Jin, M.~Guo, K.~Sreenath, X.~Lin, and P.~Abbeel, ``{RoboCopilot}: Human-in-the-loop interactive imitation learning for robot manipulation,'' \emph{arXiv preprint arXiv:2503.07771}, 2025. [Online]. Available: \url{https://arxiv.org/abs/2503.07771}
\BIBentrySTDinterwordspacing

\bibitem{xu2025crdagger}
\BIBentryALTinterwordspacing
X.~Xu, Y.~Hou, Z.~Liu, and S.~Song, ``Compliant residual {DAgger}: Improving real-world contact-rich manipulation with human corrections,'' in \emph{Advances in Neural Information Processing Systems}, 2025. [Online]. Available: \url{https://openreview.net/forum?id=cjcm5LYVWm}
\BIBentrySTDinterwordspacing

\bibitem{mandlekar2020iwr}
\BIBentryALTinterwordspacing
A.~Mandlekar, D.~Xu, R.~Mart{\'i}n-Mart{\'i}n, Y.~Zhu, L.~Fei-Fei, and S.~Savarese, ``Human-in-the-loop imitation learning using remote teleoperation,'' \emph{arXiv preprint arXiv:2012.06733}, 2020. [Online]. Available: \url{https://arxiv.org/abs/2012.06733}
\BIBentrySTDinterwordspacing

\bibitem{liu2023sirius}
\BIBentryALTinterwordspacing
H.~Liu, S.~Nasiriany, L.~Zhang, Z.~Bao, and Y.~Zhu, ``Robot learning on the job: Human-in-the-loop autonomy and learning during deployment,'' in \emph{Robotics: Science and Systems}, 2023. [Online]. Available: \url{https://ut-austin-rpl.github.io/sirius/}
\BIBentrySTDinterwordspacing

\bibitem{xia2025apo}
\BIBentryALTinterwordspacing
W.~Xia, Y.~Yang, H.~Wu, X.~Ma, T.~Kong, and D.~Hu, ``Human-assisted robotic policy refinement via action preference optimization,'' in \emph{Advances in Neural Information Processing Systems}, 2025. [Online]. Available: \url{https://arxiv.org/abs/2506.07127}
\BIBentrySTDinterwordspacing

\bibitem{pan2026sop}
\BIBentryALTinterwordspacing
M.~Pan, S.~Feng, Q.~Zhang, X.~Li, J.~Song, C.~Qu, Y.~Wang, C.~Li, Z.~Xiong, Z.~Chen, Y.~Liu, and J.~Luo, ``{SOP}: A scalable online post-training system for vision-language-action models,'' \emph{arXiv preprint arXiv:2601.03044}, 2026. [Online]. Available: \url{https://arxiv.org/abs/2601.03044}
\BIBentrySTDinterwordspacing

\end{thebibliography}
\endgroup
\end{document}